# Hybrid Approaches for our Participation to the n2c2 Challenge on Cohort Selection for Clinical Trials

Xavier Tannier[1], Nicolas Paris[2,3], Hugo Cisneros[4], Christel Daniel[1,2], Matthieu Doutreligne[2,3], Catherine Duclos[5,6], Nicolas Griffon[1,2], Claire Hassen-Khodja[7], Ivan Lerner[2,8], Adrien Parrot[2], Éric Sadou[1,2], Cyrina Saussol[2], Pascal Vaillant[5]

[1] Sorbonne Université, Inserm, Univ Paris 13, Laboratoire d'Informatique Médicale et d'Ingénierie des Connaissances pour la e-Santé, LIMICS, F-93017 Bobigny, France

[2] AP-HP, DSI-WIND, Paris, France

[3] LIMSI, CNRS, Univ. Paris-Sud, Université Paris-Saclay, F-91405, Orsay, France

[4] Mines ParisTech, 75006, Paris, France

[5] Univ Paris 13, Sorbonne Université, Inserm, Laboratoire d'Informatique Médicale et d'Ingénierie des Connaissances pour la e-Santé, LIMICS, F-93017 Bobigny, France

[6] AP-HP, Hôpital Avicenne, Unité d'Information Médicale, Bobigny, France

[7] AP-HP, Clinical Research and Innovation Department, Paris, France

[8] Université Paris Descartes - Sorbonnes Paris Cité, Paris, France

## ABSTRACT

Objective: Natural language processing can help minimize human intervention in identifying patients meeting eligibility criteria for clinical trials, but there is still a long way to go to obtain a general and systematic approach that is useful for researchers. We describe two methods taking a step in this direction and present their results obtained during the n2c2 challenge on cohort selection for clinical trials.

Materials and Methods: The first method is a weakly supervised method using an unlabeled corpus (MIMIC) to build a silver standard, by producing semi-automatically a small and very precise set of rules to detect some samples of positive and negative patients. This silver standard is then used to train a traditional supervised model. The second method is a terminology-based approach where a medical expert selects the appropriate concepts, and a procedure is defined to search the terms and check the structural or temporal constraints.

Results: On the n2c2 dataset containing annotated data about 13 selection criteria on 288 patients, we obtained an overall F1-measure of 0.8969, which is the third best result out of 45 participant teams, with no statistically significant difference with the best-ranked team.

Discussion: Both approaches obtained very encouraging results and apply to different types of criteria. The weakly supervised method requires explicit descriptions of positive and negative examples in some reports. The terminology-based method is very efficient when medical concepts carry most of the relevant information.

Conclusion: It is unlikely that much more annotated data will be soon available for the task of identifying a wide range of patient phenotypes. One must focus on weakly or non-supervised learning methods using both structured and unstructured data and relying on a comprehensive representation of the patients.

## 1. INTRODUCTION

Identifying patients sharing common clinical or biological characteristics (phenotyping) is of great importance in medical research [1, 2]. Phenotyping in electronic health patient records (EHRs) offers the possibility of phenome-wide association scans (PheWAS) for disease-gene associations [3, 4, 5]. Phenotyping plays also an important role in feasibility study and patient recruitment in clinical trials. The process of eligibility determination is extremely challenging and time-consuming, mandating most of the time manual chart review. Clinical Trial Recruitment Support Systems aim to automate this process in healthcare settings. Initially based on non-scalable, system- or vendor-specific efforts, eligibility determination is now provided by large scale platforms querying patient data from federated EHR systems from many different sites in different countries [6, 7, 8]. Numerous work has been carried out to formalize and classify eligibility criteria in order to automate patient eligibility determination [9, 10, 11, 12]. The Phenotype KnowledgeBase database contains rule-based phenotype definitions incorporating coded concepts (using SNOMED, LOINC, ICD10, etc.) [13].

However, rule-based definitions based on heterogeneous structured data such as diagnosis codes, laboratory results, medication orders or problem lists are difficult to share across different EHR systems. Moreover, the criteria, specified in natural language, can be of an extremely different nature, and some criteria require very specific medical knowledge, elaborate reasoning or subjective judgment. The notes, that still remain the preferred means of documentation for physicians, frequently contain personal or clinical information that the structured data does not, and that are critical for eligibility determination [14]. Thus, the use of clinical notes from the EHR is imperative for eligibility determination [15, 16], but there is still no effective and general approach for identifying patients who validate multiple and varied selection criteria from the text. It is also impossible to build annotated datasets for all possible criteria, which disqualifies the traditional supervised learning approaches. Early work has applied partially supervised methods using the text of the reports to try to generalize the process [17, 18, 19].

In this paper, we present two methods for minimizing the human intervention in the cohort selection: an approach using weak supervision to build a big "silver standard" dataset suitable for supervised learning, and a terminology-based approach where a medical expert selects the appropriate concepts. This work has been done in the framework of the n2c2 challenge on cohort selection for clinical trials[1] and the results presented here are the official results obtained for this challenge. This challenge aimed at investigating how the state of the art of natural language processing could be used to identify patients meeting selection criteria for clinical trials, by using only the text of the EHRs. The organizers chose 13 varied and realistic selection criteria, and the task was to design a system determining, from several textual reports on a given patient, whether this patient was eligible or not to each of these 13 criteria. More details on the criteria and the task are given at Section 3.1.

## 2. RELATED WORK

Many statistical approaches to natural language processing (NLP) have recently been proposed to exploit the texts of patient records to allow the identification of phenotypes [20, 21, 22, 23, 24, 25, 26, 27]. Most of these methods fall under the category of fully supervised learning. These models require the creation of a manually annotated dataset (by human experts) to allow the training of a statistical model, which can then be applied to new data. These approaches therefore require a considerable initial investment, and are difficult to generalize because the manual annotations are specific to a particular case [28, 25]; this is the reason why they have only been applied to a relatively limited number of phenotypes.

More recently, "weakly supervised" approaches have been proposed, making it possible to partially overcome the manual annotation step by building semi-automatically a larger dataset, but generally noisy or of lower quality. Weak supervision is a generic term for a set of approaches where labeled data is obtained by means other than full manual annotation. One of these means is "active" learning, in which the files proposed to the expert are automatically selected so as to minimize the number of patients to explore before obtaining a good quality training dataset [28]. Other methods imply a semi-automatic annotation of an unlabeled corpus with predefined terms or rules [17, 18, 19], which can be enriched with a bootstrapping algorithm. These latter works are the closest to what we propose in Section 4.2. To the best of our knowledge, other weakly supervised paradigms such as distant supervision [29] or knowledge transfer [30] have not yet been applied specifically to classifying eligibility criteria.

## 3. MATERIAL

This section describes the n2c2 corpus as well as other resources that we used for building our different systems.

### 3.1 n2c2 Corpus for Cohort Selection for Clinical Trials

We used the corpus from the n2c2 challenge on cohort selection for clinical trials, for development and evaluation. This corpus has been made available to the 45 participant teams of the challenge in 2018, and is composed of 1,267 reports in English, concerning 288 patients (between 2 and 5 documents per patient). Only the textual content of the reports is available. All patients have diabetes, most of them are at risk for heart disease. Each patient record is annotated by two medical experts who assigned to this patient a "meets" or "does not meet" status for each of the following 13 criteria:

- ABDOMINAL (*History of intra-abdominal surgery, intestine resection or small bowel*

---

> *obstruction*)

- ADVANCED-CAD (*Advanced cardiovascular disease – two or more of 4 sub-criteria defined in the guidelines*)
- ALCOHOL-ABUSE (*Current alcohol use over weekly recommended limits*)
- ASP-FOR-MI (*Use of aspirin to prevent myocardial infarction*)
- CREATININE (*Serum creatinine > upper limit of normal*)
- DIET-SUPP (*Taken a dietary supplement (excluding Vitamin D) in the past 2 months*)
- DRUG-ABUSE (*Drug abuse, current or past*)
- ENGLISH (*Patient must speak English*)
- HBA1C (*Any HbA1c value between 6.5 and 9.5%*)
- KETO-1YR (*Diagnosis of ketoacidosis in the past year*)
- MAJOR-DIABETES (*Major diabetes-related complication – among 6 complications listed in the guidelines*)
- MAKES-DECISION (*Patient must make their own medical decisions*)

The overall agreement between the two annotators before adjudication was 84.9% (overlap between two annotations).

These 288 patients are split between a training set of 202 patients and a test set of 86 patients. The latter has been provided to the participants during the test phase in May 2018. Table 1 provides more details about the corpus and **Erreur ! Source du renvoi introuvable.** shows the distribution of the "met" and "not met" classes.

| | Training set | Test set | Total |
|---|---|---|---|
| Patient number | 202 | 86 | 288 |
| Document number | 890 | 377 | 1267 |
| Mean documents par patient | 4.41 | 4.38 | 4.40 |
| Token number (approx.) | 700k | 300k | 1M |

*Table 1. n2c2 corpus general statistics.*

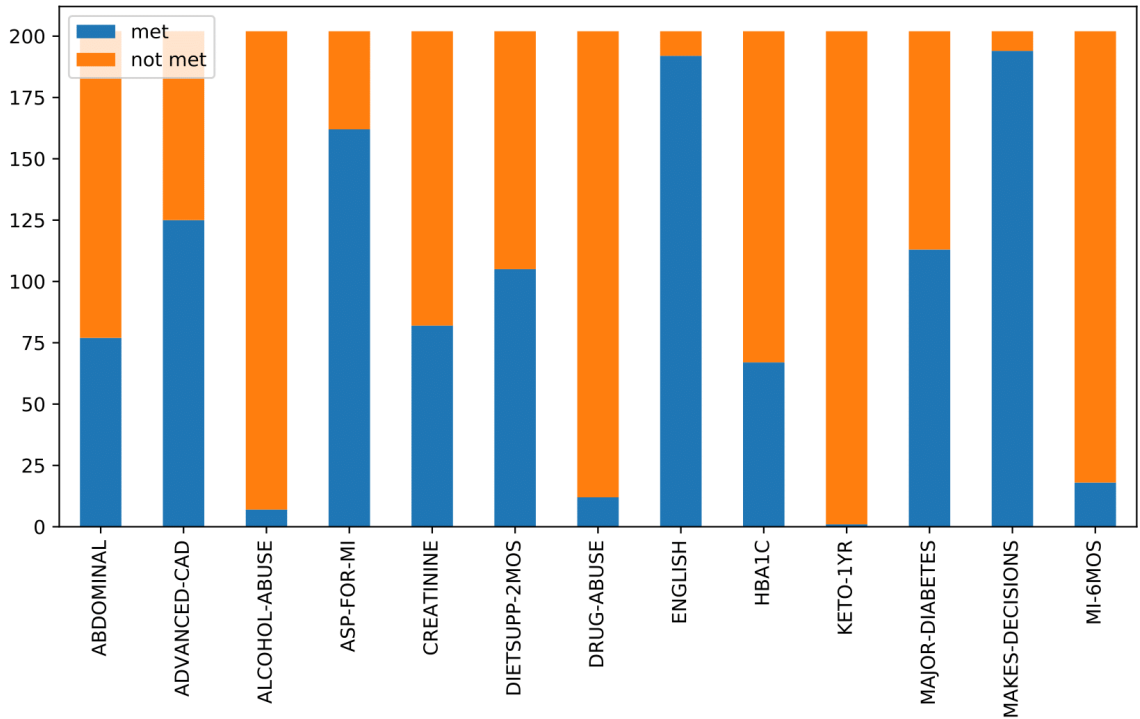

*Figure 1. Distribution of the classes "met" and "not met" for the 13 criteria of the n2c2 corpus.*

### 3.2 Other resources

- The Unified Medical Language System (**UMLS**) metathesaurus [31] comprises over 1 million biomedical concepts and 5 million concept names (mostly in English) from over 100 controlled vocabularies, including SNOMED CT which have been our main resources. We used UMLS to collect the medical concepts on interest in our terminology-based approach.

- MIMIC is a large, freely-available database comprising deidentified health-related data associated with over 40,000 patients [32]. We used the 2 million MIMIC clinical notes in English in order to compute word embeddings and augment our training set (semi-supervised approach) or find new rules (semi-supervised and rule-based approaches).

## 4. METHODS

We consider the task as 13 independent binary classification problems: given a set of several textual documents from a patient's records, and for each of the criteria described in Section 3.1, our system has to decide whether the patient meets or does not meet the criterion.

Current methods used in research for text classification are based on supervised machine learning (ML), i.e. algorithms automatically learning on the training set how to classify the patient and to generalize on unseen patients from the test set. However, some characteristics of this dataset make this approach a bad candidate for producing good results:

- Very limited amount of data available for training

- Difficult, ambiguous task requiring expert (medical) knowledge

- Highly unbalanced classes for some of the criteria (i.e. one of the two classes "met" or "not met" is very infrequent, such as ketoacidosis diagnosis or not being able to make medical decision, as illustrated by **Erreur ! Source du renvoi introuvable.**). This represents a major obstacle for most ML algorithms.

- Temporal dimension present for some criteria (e.g. ketoacidosis diagnosis *in the past year*, *current* alcohol abuse, dietary supplement *in the past 2 months*). This dimension is very hard to represent in a general representation for ML systems.

On the other hand, rule-based approaches are good candidates, but do not generalize well, which means that the 13 criteria would require 13 different hand-crafted sets of rules.

In an effort for generalizing the process of finding patients matching eligibility criteria, we investigated two methods for minimizing the human intervention: a semi-supervised approach aiming at building a big "silver standard" dataset suitable for supervised learning (Section 4.2) and a terminology-based approach where a medical expert selects the appropriate concepts (Section 4.3). Finally, some criteria were solved by a specific rule-based approach (Section 4.4). We first describe the different tools that we used to process the text (Section 4.1).

## 4.1 Text processing

We describe here the tools that we used in one or more of our systems for processing the text at different levels: tokenization, cleaning, normalization, structure extraction.

- Stanford CoreNLP [33] is a set of human language technology tools for 7 languages, not specialized in clinical texts but very robust; we used it for **sentence splitting** and **word tokenization**, in all our systems.

- Apache cTAKES [34] is an NLP system specialized in information extraction from medical record clinical free-text in English. We used it for detecting concepts that are **negated** or **uncertain** in the notes in our terminology-based approach. We also extended the negation detection with a few rules that were very frequent in our corpus.

- HeidelTime [35] is a multilingual and cross-domain **temporal tagging** tool. We used it for the few temporally constraint criteria, for detecting and normalizing dates present in the text (e.g. "*last June*" in an article written in December 2018 is converted into the absolute date "June 2018", etc.).

- **Spelling correction**. We looked for spelling mistakes and typos for all the concepts searched in our terminology-based approach. We learned a word2vec model [36] on the MIMIC clinical notes [32]. Word2vec produces vectors for each token found in the corpus; similar vectors (based here on a cosine similarity metric) correspond to words that share the same context and have a strong semantic relationship with each other (synonymy, antonyms, cohyponymy or another semantic relation). For each concept word $w$ within UMLS having more than three letters, we listed its 200 most similar words and kept those having a Levenshtein distance lower than 2 with $w$. All these words are considered as synonymous with $w$.

- **Section splitter**. Most medical documents contain some sections (e.g.: Family History, Medication...) which are often introduced by a few words ended with a newline or/and a colon. Based on this pattern we automatically extracted candidates from n2c2 and MIMIC documents and only kept most frequent (frequency > 1%). We got a total set of 1455 section titles. Based on those textual labels, we applied a regular expression pipeline to separate the documents into sections.

We now describe the different methods we used to classify the patients according to the 13 eligibility criteria.

## 4.2 Weakly supervised methods

The idea underlying this first approach is to massively augment the training data with the help of the MIMIC database, which contains more than 2 million unlabeled clinical notes for more than 40,000 patients. The three criteria concerned by this approach are:

- ALCOHOL-ABUSE (*Alcohol use over weekly recommended limits*)
- DRUG-ABUSE (*Drug abuse, current or past*)
- KETO-1YR (*Diagnosis of ketoacidosis in the past year*)

**Erreur ! Source du renvoi introuvable.** illustrates the process with the example of ALCOHOL-ABUSE. From the n2c2 dataset, we have produced semi-automatically (automatic extraction and manual review and correction) a small and very precise set of rules to detect without ambiguity some samples of positive ("met") or negative ("not met") patients (step ① in the figure). Examples of positive and negative rules for ALCOHOL-ABUSE and DRUG-ABUSE are given at Table 2; note that positive examples are discarded when found in the context of a negation ("*no history of*", "*denies*", etc.). Being very specific, these rules are expected to have a high precision but a very bad coverage (recall), because of the high linguistic variation that can be observed in the reports. However, applied to a big corpus such as MIMIC, these rules are enough to collect a large number of patients (ALCOHOL-ABUSE: 13,000 negative, 1,500 positive ; DRUG-ABUSE: 6,100 negative, 2,000 positive) (step ②). As these patients are often associated to many reports, our hypothesis is then that expressions other than the seed rules will be used for describing the same problem (alcohol, drug abuse, ketoacidosis diagnosis) through the patient's file, leading to a large and varied training corpus. Such a corpus, built semi-automatically, is called a silver standard, as opposed to a gold standard verified and validated by human experts. Similar rules have been created for DRUG-ABUSE, while we used the ICD-9 ketoacidosis diagnosis code for KETO-1YR (250.1 and children).

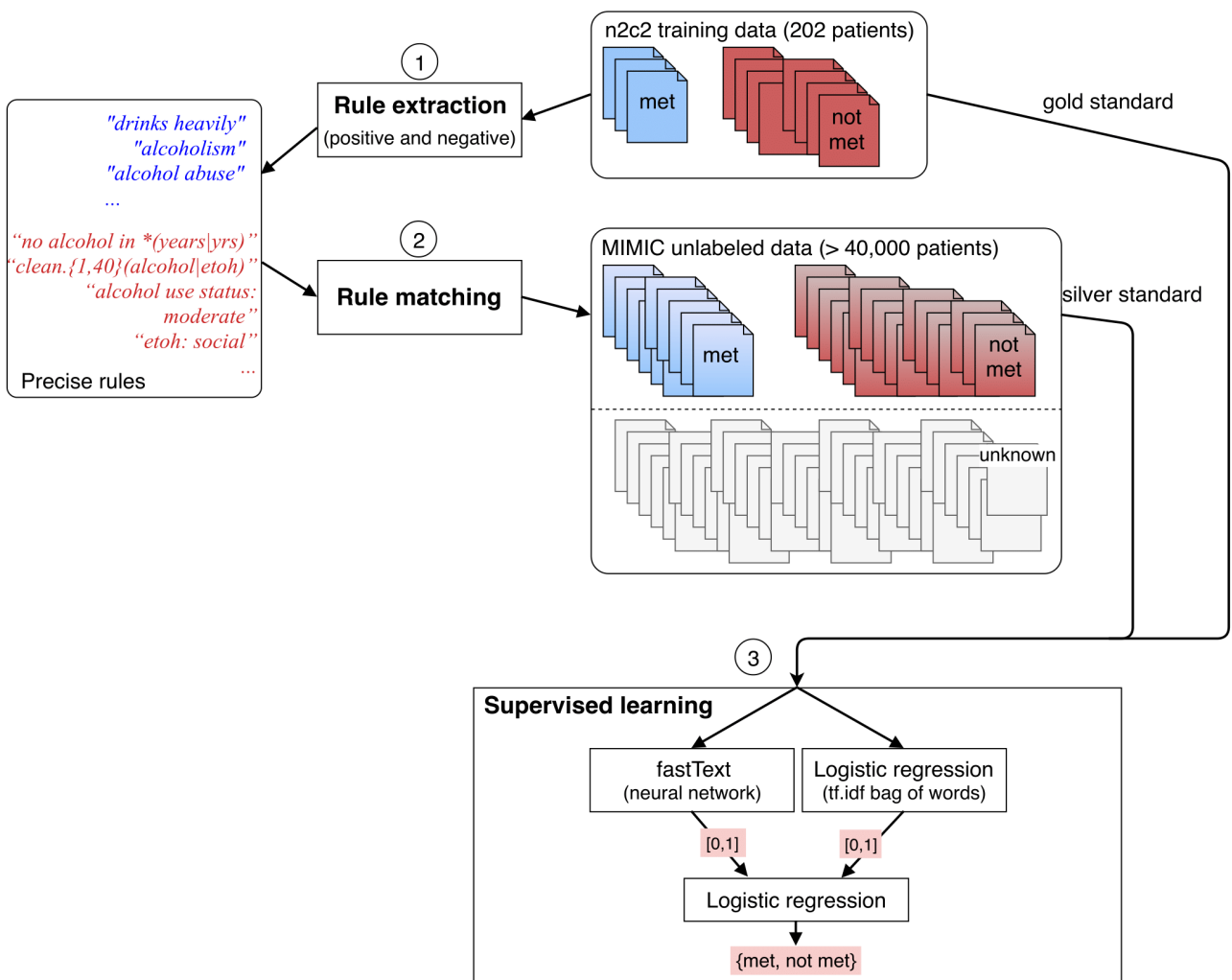

*Figure 2. Examples of trigger rules for alcohol-abuse.*

| ALCOHOL-ABUSE | |
|---|---|
| Positive rules | Negative rules |
| *"drinks heavily"*<br>*"alcoholism"*<br>*"alcohol abuse"*<br>*"binge drinking"*<br>*"drinking episodes"*<br>*"etoh"*<br>*(20 rules total)* | *"denies [...] alcoholic beverage abuse"*<br>*"no alcohol in * (years|yrs)"*<br>*"clean.{1,40} (alcohol|etoh)"*<br>*"alcohol use status: moderate"*<br>*"etoh: social"*,<br>*etc.*<br>*(40 rules total)* |
| DRUG-ABUSE | |
| Positive rules | Negative rules |
| *"uses { DRUG_NAMES }"*<br>*"used { DRUG_NAMES }"*<br>*"consumed { DRUG_NAMES }"*<br>*"previous use of { DRUG_NAMES }"*<br>*"history of { DRUG_NAMES }"*<br>*With DRUG_NAMES a set containing ["marijuana", "cocaine", "heroin", "crack", "ivdu", etc.](32 names total ~ 200 rules)* | *"never used(?={DRUG_NAMES})"*<br>*"clean.{1,40} {DRUG_NAMES}"*<br>*"no.*drug use.*history"*<br>*"negative for.{0,30} {DRUG_NAMES}"*<br>*"(?={DRUG_NAMES})(?=(son|brother|etc.))*<br>*(180 rules total)* |

*Table 2: Examples of trigger rules for* ALCOHOL-ABUSE.

This silver standard, together with the n2c2 training data (gold standard) can then be used to train a classical supervised classification system, with enough data to expect good results. In our case, for DRUG-ABUSE and ALCOHOL-ABUSE, we applied two learning models (step ③): the deep learning classifier FastText [37] using embeddings pre-trained on MIMIC, and a simple logistic regression based on a tf.idf transformation of the words. The final prediction is provided by another logistic regression on the predictions of the first two models. For KETO-1YR, which had no positive example in the training data, we choose a simpler random forest algorithm which finally correctly predicted zero positive example in the test data as well.

Note that all these fully supervised training systems need both positive and negative examples for their training. For this reason, an effective weakly-supervised approach requires that positive and negative rules can be written, which is actually quite rare, as will be noticed in next sections and discussed in Section 5.

### 4.3 Terminology-based methods

Our second effort to generalize the process of cohort selection applies to criteria for which medical knowledge is important and the presence of specific medical concepts in the texts is crucial to decide about eligibility. We identified six criteria having these characteristics:

- ABDOMINAL (*History of intra-abdominal surgery, intestine resection or small bowel obstruction*)
- ADVANCED-CAD (*Advanced cardiovascular disease)*
- ASP-FOR-MI (*Use of aspirin to prevent myocardial infarction*)
- DIET-SUPP (*Taken a dietary supplement (excluding Vitamin D) in the past 2 months*)
- MAJOR-DIABETES (*Major diabetes-related complication*)
- MI-6MOS (*Myocardial infarction in the past 6 months*)

For all these cases, negative ("not met") situations are rarely explicit in the documents: the text does almost never specify that the patient has *not* taken any dietary supplement or has had *no* myocardial infarction, and so on. That is why we were not able to apply a semi-supervised strategy as described in previous section. Instead, we applied a concept-matching approach based on the following steps:

1. **Terminology**: use UMLS to collect descendants and synonyms of concepts corresponding to the inclusion criteria, as defined by medical experts in the team.

2. **Spelling correction**: apply our word2vec-based spell checker.

3. **Negation, uncertainty**: use cTAKES [34] and our extra rules to discard negated or uncertain events in text (e.g. "*The patient is being evaluated for renal transplant*" or "*discussed possibility of gastric bypass*" for ABDOMINAL, "*she has no evidence of neuropathy at this time*" for MAJOR-DIABETES).

4. **Structure**: restrict the concept search to specific sections, as defined either by medical experts or by an automatic selection of relevant sections within the training data. For example, "*calcium*" is a dietary supplement if found in a "Medication" section, but not in a "laboratory result" section.

5. **Temporal dimension**: if specified in the description, restrict to recent reports or to sentences containing a date within the inclusion date range (with HeidelTime [35]). E.g.: "*gentleman who had an MI in June 2017*" for the class MI-6MOS.

The final step is to trigger a "met" decision as soon as at least one relevant term is found in the text.

This process requires a significant human investment, but less than an exclusively rule-based approach.

### 4.3.1    Example: ABDOMINAL

For this criterion, positive patients must have a history of intra-abdominal surgery, small or large intestine resection or small bowel obstruction. Hernia are explicitly excluded from this definition. Based on this description, we collected all descendants of the concept "Operation on abdominal region" (CUI C0198482) in the UMLS, as well as "small bowel obstruction", "stomach bypass", "hysterectomy", "cholecystectomy", and we excluded 19 CUIs containing the word "hernia", as well as noisy terms having very common homonyms in the training corpus ("apr", "turp", "tips", "sch"). This led to a list of 17,703 terms that were systematically searched for in the texts. Negated or uncertain terms were discarded. Each patient file containing at least one of these terms was labeled as "met".

### 4.3.2    Example: DIET-SUPP

For this criterion, positive patients must have taken a dietary supplement (excluding Vitamin D) in the past 2 months. We collected a list of 183 ingredients listed as dietary supplements in different resources and mapped this list to the records created between "now" (for the challenge, "now" was the date of the last report for the patient) and 2 months before. We looked only in the sections involving new medications.

### 4.3.3    Example: ADVANCED-CAD

The guidelines define "advanced" as having two or more of the following:

- Taking two or more medications to treat CAD. The CAD medication list was built from the structured prescription tables of MIMIC, as the difference between the medications prescribed for general coronary diseases (ICD9 in 410, 411, 412, 413) and the others. We manually reviewed the result and extended it with medications from NHS guidelines[2] to result in a 40 entry list.

---

[2] https://www.nwcscnsenate.nhs.uk/files/9014/5642/1864/CMSCN_NSTEMI_ACS_Guideline_Final_2016.pdf

- History of myocardial infarction. Met when "infarction", "stemi" and synonyms are present except together with other organs than heart and excluding sections specific to neuro imaging.
- Currently experiencing angina. Met when "angina" or synonyms is present with no date or together with a date in the same sentence which is less than 6 months old.
- Ischemia, past or present. Met when "ischemia" or synonyms is present.

### 4.4    Rule-based methods

Within the n2c2 dataset, 4 criteria unfortunately escape the generalization efforts and require the implementation of dedicated rules:

- CREATININE (*Serum creatinine > upper limit of normal*)
- HBA1C (*Any HbA1c value between 6.5 and 9.5%*)
- ENGLISH (*Patient must speak English*)
- MAKES-DECISION (*Patient must make their own medical decisions*)

The first two are lab results that can be extracted quite easily from tables or free text, with little variations. Each one required no more than 15 basic regular expressions. We used the annotated training data to determine the "upper limit of normal" for creatinine, finally set to 1.5 mg/dL (disregarding the male or female status of the patient), which is above the usual recommendations[3].

ENGLISH and MAKES-DECISION are much more difficult because of three main reasons:

- Data is highly unbalanced (most patients can make their own decisions and speak English in this U.S. dataset, see **Erreur ! Source du renvoi introuvable.**).
- Negative examples are implicit (the report will not mention that the patient can make their decisions or speak English), making our semi-supervised approach impossible to set up.
- There is an infinity of possible variations to express the fact that a patient is unable to speak English or make decisions (e.g. a long anecdote leading to the implicit conclusion that the patient is mentally too confused to meet the criterion).

For these reasons, rules currently seem to be the only way to go. We built a set of weighted positive and negative rules around the idea that a patient could speak another language, need a translator or not understand English (ENGLISH, 55 rules), or have dementia, mental retardation, be unable to answer questions or under tutorship or curatorship (MAKES-DECISION, 67 rules).

## 5.    RESULTS

During the n2c2 challenge, our different systems have been evaluated by the organizers on the 86 patients from the test set, using the traditional metrics precision, recall and F1-measure (harmonic mean of precision and recall). Table 3 reproduces these results, together with the support for each class (number of "met"/"not met" patients). Following the official global metric of the challenge (micro F1-measure of 0.8969), we achieved the third best result out of 45 participant teams, with no statistically significant difference[4] with the best-ranked team (Medical University of Graz –  F1 0.9100) that applied a dedicated rule-based system to each criterion.

---

[3] 1.0 mg/dL for females and 1.2 mg/dL for males.

[4] As reported by the organizers.

| Approach | Class / Criterion | Met | | | | Not met | | | | Overall F1 |
|---|---|---|---|---|---|---|---|---|---|---|
| | | *Sup.* | Prec. | Rec. | F1 | *Sup.* | Prec. | Rec. | F1 | |
| Semi-supervised | ALCOHOL | *3* | 1.00 | 0.33 | 0.50 | *83* | 0.98 | 1.00 | 0.99 | 0.74 |
| | DRUG | *3* | 0.67 | 0.67 | 0.67 | *83* | 0.99 | 0.99 | 0.99 | 0.83 |
| | KETO-1YR | *0* | -[5] | -[5] | -[5] | *86* | 1.00 | 1.00 | 1.00 | 1.00[5] |
| Terminology-based | ABDOMINAL | *30* | 0.85 | 0.93 | 0.89 | *56* | 0.96 | 0.91 | 0.94 | 0.91 |
| | ADV.-CAD | *45* | 0.74 | 0.87 | 0.80 | *41* | 0.82 | 0.66 | 0.73 | 0.76 |
| | ASP-FOR-MI | *68* | 0.83 | 0.91 | 0.87 | *18* | 0.45 | 0.28 | 0.34 | 0.61 |
| | DIETSUPP | *44* | 0.91 | 0.89 | 0.90 | *42* | 0.88 | 0.90 | 0.89 | 0.90 |
| | M.-DIABETES | *43* | 0.90 | 0.86 | 0.88 | *43* | 0.87 | 0.91 | 0.89 | 0.88 |
| | MI-6MOS | *8* | 0.67 | 0.50 | 0.57 | *78* | 0.95 | 0.97 | 0.96 | 0.77 |
| Rule-based | CREATININE | *24* | 0.87 | 0.83 | 0.85 | *62* | 0.93 | 0.95 | 0.94 | 0.90 |
| | HBA1C | *35* | 1.00 | 0.77 | 0.87 | *51* | 0.86 | 1.00 | 0.93 | 0.90 |
| | ENGLISH | *73* | 0.94 | 0.99 | 0.96 | *13* | 0.89 | 0.62 | 0.73 | 0.84 |
| | M.-DECISION | *83* | 0.96 | 0.98 | 0.97 | *3* | 0.00 | 0.00 | 0.00 | 0.49 |
| | overall (micro) | - | 0.88 | 0.90 | 0.89 | - | 0.93 | 0.92 | 0.92 | 0.9069 |
| | overall (macro) | - | 0.79 | 0.73 | 0.75 | - | 0.81 | 0.78 | 0.79 | 0.8095[5] |

*Table 3. Our official results at n2c2 challenge.*

## 6. DISCUSSION

The scores presented at previous section are encouraging, even though one must be very careful when interpreting results for some criteria supported by only few positive or negative examples. The terminology-based approach obtained very good results, except for ASP-FOR-MI where some more reasoning is probably necessary. The weakly-supervised approach led to coherent results, showing that this lead is worth being followed. Its limits should also be further investigated: first, we must work on evaluating and reducing the noise and possible bias introduced by the method used to collect the silver standard. Second, we pointed out that finding negative examples was the trickiest issue with this approach. In previous work, Agarwal et al. [18] build a silver standard in which positive and negative examples are documents where some predefined terms are respectively present and absent. Halpern et al. [17] find only positive examples and apply Positive-Unlabeled learning algorithm [38] to build their silver standard. These approaches would generalize to more cases but generate noisier datasets.

It is therefore notable that the different existing approaches apply to different categories of criteria, which must be identified in advance: a terminological approach (Section 4.3) when only medical terms are manipulated and the presence of or the absence of certain concepts is generally sufficient to discriminate between positive and negative examples; a weakly supervised approach seeking positive and negative examples (Section 4.2) where possible, that is, both classes are sometimes made explicit in the reports; a weakly supervised approach like those of Agarwal et al. [18] or Halpern et al. [17] when the negative examples are implicit.

---

[5] Note that the official evaluation script attributed the erroneous precision and recall of 0 for "met" on KETO-1YR, due to the fact that there was no "met" in the test set. This value is then different than the official one.

Finally, note that the metrics used during the challenge and this article (especially F1-measure) mask that fact that a very high recall is expected from these systems, in order to guarantee that only few patients are missed and that the system can be validated as an efficient filter before human intervention. Rule-based systems are generally more precise than sensitive.

## 7. CONCLUSION AND PERSPECTIVES

There is still a way to go to get tools able to identify a wide range of patient phenotypes for cohort selection, or even to help the researcher build cohorts faster and with less potential bias. In this paper, we have outlined approaches for the generalization of the process, with results as good as fully hand-crafted rule-based systems.

Building the n2c2 dataset used in our work has been a huge effort, and the organizers must be thanked for this. However, we cannot expect more data on many more criteria to be produced in the near future. The effort must now focus on creating weakly or non-supervised methods able to actively assist the medical researchers, to transfer a model from a phenotype to another or to learn generic patient representations.

Moreover, next generation phenotyping requires methods allowing the joint exploitation of structured and unstructured data. Using all available information and knowledge seems to be a necessary condition for getting closer to the performance of a human expert for patient selection. The exploitation of such heterogeneous data is a challenge, but recent advances in the field of information representation, notably thanks to neural networks, show that all types of structure can be represented in the same space, allowing the implementation of algorithms on a single mode of representation from multiple sources. This type of approaches has been applied, for example, to the joint representation of images and text [39], knowledge bases and texts [40], and very recently patients with structured data and text [41].